\documentclass[a4paper,fleqn]{cas-dc}

\def\tsc#1{\csdef{#1}{\textsc{\lowercase{#1}}\xspace}}
\tsc{WGM}
\tsc{QE}
\tsc{EP}
\tsc{PMS}
\tsc{BEC}
\tsc{DE}

\usepackage{graphicx}  
\usepackage{caption}
\usepackage{float}
\usepackage{stfloats}
\usepackage{amsmath}
\usepackage{booktabs}
\usepackage{pifont}
\usepackage{soul}
\usepackage[numbers,sort&compress]{natbib}
\usepackage{multirow}
\usepackage{subfigure} 
\usepackage{ulem}
\usepackage{algorithm}
\usepackage{listings}
\begin{document}
\let\WriteBookmarks\relax
\def\floatpagepagefraction{1}
\def\textpagefraction{.001}
\let\printorcid\relax   
\shorttitle{Image and Vision Computing} 
\setlength{\abovedisplayskip}{2pt}


\title [mode = title]{LDConv: Linear deformable convolution for improving convolutional neural networks}                      



%

\author[1]{Xin Zhang}
\author[1]{Yingze Song}
\author[1]{Tingting Song}
\cormark[1]
\ead{E-mail address: ttsong@cqnu.edu.cn}
\author[1,2]{Degang Yang}
\author[3]{Yichen Ye}
\author[1]{Jie Zhou}
\author[1]{Liming Zhang}






\affiliation[1]{organization={College of Computer and Information Science, Chongqing Normal University},
	city={Chongqing},
	postcode={401331}, 
	country={china}}



\affiliation[2]{organization={Chongqing Engineering Research Center of Educational Big Data Intelligent Perception and Application},
  city={Chongqing},
  postcode={401331}, 
  country={china}}

\affiliation[3]{organization={College of Electronic and Information Engineering, Southwest University},
	city={Chongqing},
	postcode={400715}, 
	country={china}}

\cortext[cor1]{Corresponding author.}




\begin{abstract}
Neural networks based on convolutional operations have achieved remarkable results in the field of deep learning, but there are two inherent flaws in standard convolutional operations. On the one hand, the convolution operation is confined to a local window, so it cannot capture information from other locations, and its sampled shapes is fixed. On the other hand, the size of the convolutional kernel are fixed to k $\times$ k, which is a fixed square shape, and the number of parameters tends to grow squarely with size. Although Deformable Convolution (Deformable Conv) address the problem of fixed sampling of standard convolutions, the number of parameters also tends to grow in a squared manner, and Deformable Conv do not explore the effect of different initial sample shapes on network performance. In response to the above questions, the Linear Deformable Convolution (LDConv) is explored in this work, which gives the convolution kernel an arbitrary number of parameters and arbitrary sampled shapes to provide richer options for the trade-off between network overhead and performance. In LDConv, a novel coordinate generation algorithm is defined to generate different initial sampled positions for convolutional kernels of arbitrary size. To adapt to changing targets, offsets are introduced to adjust the shape of the samples at each position. LDConv corrects the growth trend of the number of parameters for standard convolution and Deformable Conv to a linear growth. Compared to Deformable Conv, LDConv provides richer choices and can be equivalent to deformable convolution when the number of parameters of LDConv is set to the square of K. Differently, this paper also explores the effect of neural networks by using LDConv with the same size and different initial sampling shapes. LDConv completes the process of efficient feature extraction by irregular convolutional operations and brings more exploration options for convolutional sampled shapes. Object detection experiments on representative datasets COCO2017, VOC 7+12, and VisDrone-DET2021 fully demonstrate the advantages of LDConv. LDConv is a plug-and-play convolutional operation that can replace the convolutional operation to improve network performance. The code for the relevant tasks can be found at \url{https://github.com/CV-ZhangXin/LDConv}.
\end{abstract}


\maketitle

\section{Introduction}
Convolutional Neural Networks (CNNs), such as ResNet \cite{he2016deep}, DenseNet \cite{huang2017densely}, and YOLO \cite{redmon2016you}, have demonstrated excellent performance in various applications and have led the technological progress in many aspects of modern society. It has become indispensable, from image recognition in self-driving cars \cite{chang2022yolo} and medical image analysis \cite{xie2021cotr} to intelligent surveillance \cite{abbasi2021deep} and personalized recommendation systems \cite{an2022design}. These successful networks heavily rely on convolutional operations, which efficiently extract local features in images and ensure model complexity.

Despite the fact that CNNs have achieved many successes in classification \cite{qin2020biological}, object detection \cite{wang2023improved}, semantic segmentation \cite{yang2023drnet}, etc., they still have some limitations. One of the most notable limitations concerns the choice of convolutional sampled shaped and size. Standard convolution operations tend to rely on square kernels with fixed sampled locations, such as 1 $\times$ 1, 3 $\times$ 3, 5 $\times$ 5, and 7 $\times$ 7, etc. The sampled position of the conventional kernel is not deformable and cannot be 
transformed dynamically in response to changes in various objects. Deformable Conv \cite{dai2017deformable, zhu2019deformable} enhances network performance by using the offset to flexibly adjust the sampled shape of the convolution kernel, which adapts to the change of the target. For instance, in \cite{zhao2022attentional,song2023lightweight,huang2021fapn}, these works utilized it to align features. Zhao et al. \cite{zhao2022lightweight} enhanced the detection performance of dead fish by adding it to YOLOv4 \cite{bochkovskiy2020yolov4}. Yang et al. \cite{yang2023deformable} improved the YOLOv8 \cite{YOLOV8} for cattle detection by incorporating it into the backbone. Li et al. \cite{li2022deep} introduced Deformable Conv into deep image compression tasks \cite{dumas2019context,balle2018variational} to obtain content-adaptive receptive-fields. Although the studies mentioned above have demonstrated the superior benefits of Deformable Conv, it is still not flexible enough. Deformable Conv can only define k $\times$ k convolution operations to extract features, i.e. the number of parameters is 1, 4, 9, 16, 25, ..., and so on. In fact, Deformable Conv requires resampling of features by coordinates and then extracting features by regular convolution operations. Therefore, the number of parameters of Deformable Conv tends to increase by a square, which makes the computational and memory overhead of the resampling and convolutional feature extraction processes difficult to regulate. When processing images, we may need to apply larger size deformable convolution to extract information. For example, the size of 5 $\times$ 5 deformable convolution is adjusted to a large 6 $\times$ 6 convolution kernel, i.e., the number of parameters is directly transformed from 25 to 36. This abrupt change may lead to insufficient memory on the device. Therefore, the trend of squared growth of the number of parameters in Deformable Conv lacks flexibility in setting the convolution size, and is not a friendly way of growing in the hardware environment, which ignores convolutions with parameter numbers 3, 5, 7, 8, 10, 11, 12, and so on. Moreover, Deformable Conv has not explored the effect of different initial sampling shapes on the network.

In response to the above question, we propose the flexible Linear Deformabole Conv (LDConv), which is able to flexibly adjust the size of the number of convolutional kernel parameters to adapt to changes in the shape of the object.
Unlike standard regular convolution, LDConv is a novel convolutional operation, which can extract features using efficient convolution kernels with any number of parameters, such as (1, 2, 3, 4, 5, 6, 7, etc.), whereas it is not implemented in standard convolution or Deformable Convolution. LDConv can easily be used to replace the standard convolutional operations in the network to improve network performance. Importantly, LDConv allows the number of convolutional parameters to trend linearly up or down, which is beneficial to hardware environments, and it can be used as an alternative to lightweight models to reduce the number of parameters and computational overhead. Secondly, it has more options for larger-size kernels to enhance the performance of the network when resources are sufficient. The regular convolutional kernel makes the number of parameters show a square increasing trend, while LDConv only shows a linear increasing trend and provides more options for the convolution kernel. Furthermore, its ideas can be extended to specific areas because the specially sampled shapes can be created for convolutional operations according to prior knowledge, and then dynamically and automatically adapt to changes in the target shape via offsets. Meanwhile, LDConv can easily be added to some novel convolutional modules to enhance its performance, such as FasterBlock \cite{chen2023run} and GSBottleneck \cite{li2022slim}. Object detection experiments on representative datasets VOC \cite{everingham2015pascal}, COCO2017 \cite{lin2014microsoft}, VisDrone-DET2021 \cite{zhu2021detection} fully demonstrate the advantages of LDConv. In summary, our contributions are as follows:

\begin{itemize}
	
	\item For different sizes of convolutional kernels, an algorithm is proposed to generate initial sampled coordinates for convolutional kernels of arbitrary sizes.
	
	\item To adapt to the different variations of the targets, the sampled position of the irregular convolutional kernel is adjusted by the obtained offsets. Meanwhile, three methods are also explored to extract features corresponding to irregular convolutional kernels.
	
	\item Compared to regular convolution kernels, the proposed LDConv realizes the function of irregular convolution kernels to extract features, providing convolution kernels with arbitrary sampled shapes and sizes for a variety of varying targets, which makes up for the shortcomings of regular convolutions.
	
	\item Corresponding sizes of LDConv are used to replace the convolutions in FasterBlock and GSBottleneck to improve the performance of both modules. 
\end{itemize}

\section{Related works}
In recent years, many works have considered and analyzed standard convolutional operations from different perspectives, and then designed novel convolutional operations to improve network performance.

Li et al. \cite{li2021involution} argued that convolutional kernels share parameters across all spatial locations, which leads to limited modelling capabilities across different spatial locations, and does not effectively capture spatial long-range relationships. Secondly, the approach of using a different convolution kernel for each output channel is actually not efficient. Therefore, to address these shortcomings, they proposed the Involution operator, which inverts the features of the convolutional operation to improve network performance. Qi et al. \cite{qi2023dynamic} proposed the DSConv based on Deformable Conv. The offset obtained from learning in Deformable Conv is free, which leads to the model losing a small percentage of fine structure features. This poses a great challenge for the task of segmenting elongated tubular structures, therefore, they proposed the DSConv. Zhang et al. \cite{zhang2023rfaconv} understood the spatial attention mechanism form a new perspective, they asserted that the spatial attention mechanism essentially solves the problem of parameter sharing of convolutional operations. However, some spatial attention mechanisms, such as CBAM \cite{woo2018cbam} and CA \cite{hou2021coordinate}, do not completely solve the problem of large-size convolutional parameter sharing. Therefore, they proposed RFAConv. Chen et al. \cite{chen2020dynamic} proposed the Dynamic Conv. Unlike using a convolutional kernel for each layer, the Dynamic Conv dynamically aggregated multiple parallel convolutional kernels based on their attention. The Dynamic Conv provided a greater representation of features.
Tan et al. \cite{tan2019mixconv} argued that kernel size is often neglected in CNNS, which may affect the accuracy and efficiency of the network. Second, using only layer-by-layer convolution does not utilize the full potential of convolutional networks. Therefore, they proposed MixConv, which naturally mixes multiple kernel sizes in a single convolution to improve the performance of networks.

To address the limitations in traditional sequence data processing, Romero et al.\cite{DBLP:conf/iclr/RomeroKBTH22} proposed CKConv, which overcomes the shortcomings of CNNs by treating the convolution kernel as a continuous function rather than a series of independent weights. CKConv can define arbitrarily large memory ranges in a single operation, independent of network depth, expansion factor, or network size. When designing a convolutional neural network, choosing the right convolutional kernel size is critical to model performance. Traditional methods require fixing the kernel size before training, but learning the kernel size dynamically during the training process is beneficial to improving network performance. Therefore, Romero et al. \cite{flexConv} proposed FlexConv, which is able to learn high bandwidth convolutional kernels with variable sizes during training while maintaining a fixed number of parameters. In CNN, the receptive-field can be expanded simply by increasing the size of the convolutional kernel.  Yet the number of trainable parameters, which scales quadratically with	the kernel’s size in the 2D case, rapidly becomes prohibitive, and the training is	notoriously difficult. Therefore, to use the large receptive field for convolutional neural networks and to avoid the number of parameters and computational costs, Hassani et al. \cite{hassani2023dilated} proposed the DCLS method, which can achieve a larger receptive field while keeping the number of parameters constant by controlling two hyperparameters: the number of kernels and the size of the expanded kernel. The setting of the convolutional kernel size in CNN networks is restricted, and it is usually necessary to pre-set the convolutional kernel size, making the tuning of hyperparameters cumbersome. Therefore, Pintea et al.\cite{pintea2021resolution} proposed N-JetNet, which utilizes scale space theory to obtain a self-similar parameterisation of the convolutional kernel.

Although these methods improve the performance of convolutional operations, some works are still limited to regular convolutional operations and do not allow multiple variations of convolutional sampled shapes. Other works allow for flexible resizing of the convolution kernel, but the performance obtained by the network is still less than optimal. In contrast, our proposed LDConv can efficiently extract features using a convolutional kernel with an arbitrary number of parameters and sampled shapes to realize good performance for networks.

\section{Methods}
\subsection{Define the initial sampled position}
Convolutional neural networks, based on the convolution operation, localize the features at the corresponding locations by means of a regular sampled grid. In \cite{zhao2018distortion,coors2018spherenet,dai2017deformable}, the regular sampled grid for the 3 × 3 convolution operation is given. Let R denote the sampled grid, then R is denoted as follows:
\begin{equation}
\begin{aligned}
R = \left \{(-1,-1), (-1,0),..., (0, 1), (1, 1)  \right  \}
\end{aligned}
\end{equation}

However, since the sampled coordinates defined in Deformable Conv and these works are regular, the process of efficiently extracting features by irregular convolution can not be accomplished. While LDConv targets irregularly shaped convolutional kernels. Therefore, to allow irregular convolutional kernels to have a sampled grid, an algorithm is created for arbitrary-size convolution, which generates the initial sampled coordinates $P_{n}$ of the convolutional kernel. The regular sampled grids are generated firstly, then the irregular sampled grids are created for the remaining sampled points, and finally they are stitched together to form the overall sampled grid. The pseudo code is as in Algorithm~\ref{alg:code}.

\begin{algorithm}[h]
	\caption{\small{Pseudo-code for initial coordinate generation for the convolution kernel in a PyTorch-like.}}
	\label{alg:code}
	\definecolor{codeblue}{rgb}{0.25,0.5,0.5}
	\lstset{
		backgroundcolor=\color{white},
		basicstyle=\fontsize{7.2pt}{7.2pt}\ttfamily\selectfont,
		columns=fullflexible,
		breaklines=true,
		captionpos=b,
		commentstyle=\fontsize{7.2pt}{7.2pt}\color{codeblue},
		keywordstyle=\fontsize{7.2pt}{7.2pt},
	}
	\vskip -0.075in
	\begin{lstlisting}[language=python]
	# func get_p_n(num_param, dtype)
	# num_param: the kernel size of LDConv
	# dtype: the type of data
	
	####### function  body  ########
	# get a base integer to define coordinate
	base_int = round(math.sqrt(num_param))
	row_number = num_param // base_int
	mod_numer = num_param % base_int
	
	# get the sampled coordinate of regular kernels
	
	p_n_x,p_n_y = torch.meshigrid(
	torch.meshgrid(0, row_numb)
	torch.meshgird(0, base_int))
	
	# flatten the sampled coordinate of regular kernels
	p_n_x = torch.flatten(p_n_x)
	P_n_y =  torch.flatten(p_n_y)
	
	# get the sampled coordinate of irregular kernels
	If mod_number > 0:
		mod_p_n_x, mod_p_n_y = torch.meshgird(
		torch.arange(row_number, row_number + 1),
		torch.arange(0, mod_number))
	
		mod_p_n_x = torch.flatten(mod_p_n_x)
		mod_p_n_y = torch.flatten(mod_p_n_y)
		P_n_x,p_n_y = torch.cat((p_n_x,mod_p_n_x)),torch.cat((p_n_y,mod_p_n_y))
	
	# get the completed sampled coordinate
	p_n = torch.cat([p_n_x, p_n_y], 0)
	p_n = p_n.view(1, 2 * num_param, 1, 1).type(dtype)
	return p_n
	\end{lstlisting}
	\vskip -0.075in
\end{algorithm}

\begin{figure}[h]
	\centering
	\includegraphics[trim=0 0 0 0,clip,scale=1]{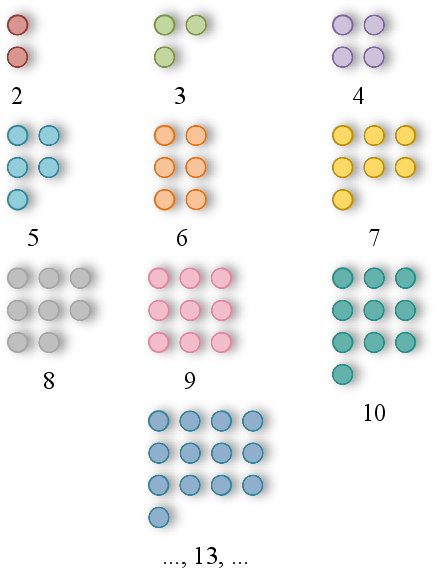}
	\caption{ The initial sampled coordinates for arbitrary convolutional kernel sizes are generated by a generation algorithm. It provides initial sampled shapes for irregular convolution kernel sizes.}
	\label{Kernels-samples}
\end{figure}

As shown in Figure~\ref{Kernels-samples}, initial sampled coordinates are generated for arbitrary size convolution. The setting of the initial sampling shapes of these convolutional kernels with different numbers of parameters needs to be considered in two ways. Firstly, different initial sampling shapes can affect the performance of the network in the same size case, while approximating a square shape facilitates Offset learning, as shown in the subsequent exploratory experiments in Section 4.5. Second, to facilitate the generation of arbitrary convolutional kernel sizes. Therefore, the sampling shape in Figure~\ref{Kernels-samples} considers the above two points to generate the corresponding initial sampling shape for the convolution of arbitrary size. The sampled grid for regular convolution treats the center as the origin (0, 0), while most of the irregular convolutions are not centered. To adapt the size of the convolution used, the point in the upper left corner is set as the sampled origin (0, 0) in the algorithm.

After defining the initial coordinates $P_{n}$ for the irregular convolution, the corresponding convolution operation at position $P_{0}$ can be defined as follows:
\begin{equation}
\begin{aligned}
\operatorname{Conv}\left(P_0\right)=\sum_{P_{n} \in R} w_{p_n} \times x\left(P_o+P_n\right)
\end{aligned}
\end{equation}

\begin{figure*}[h]
	\centering
	\includegraphics[trim=0 0 0 0,clip,scale=0.58]{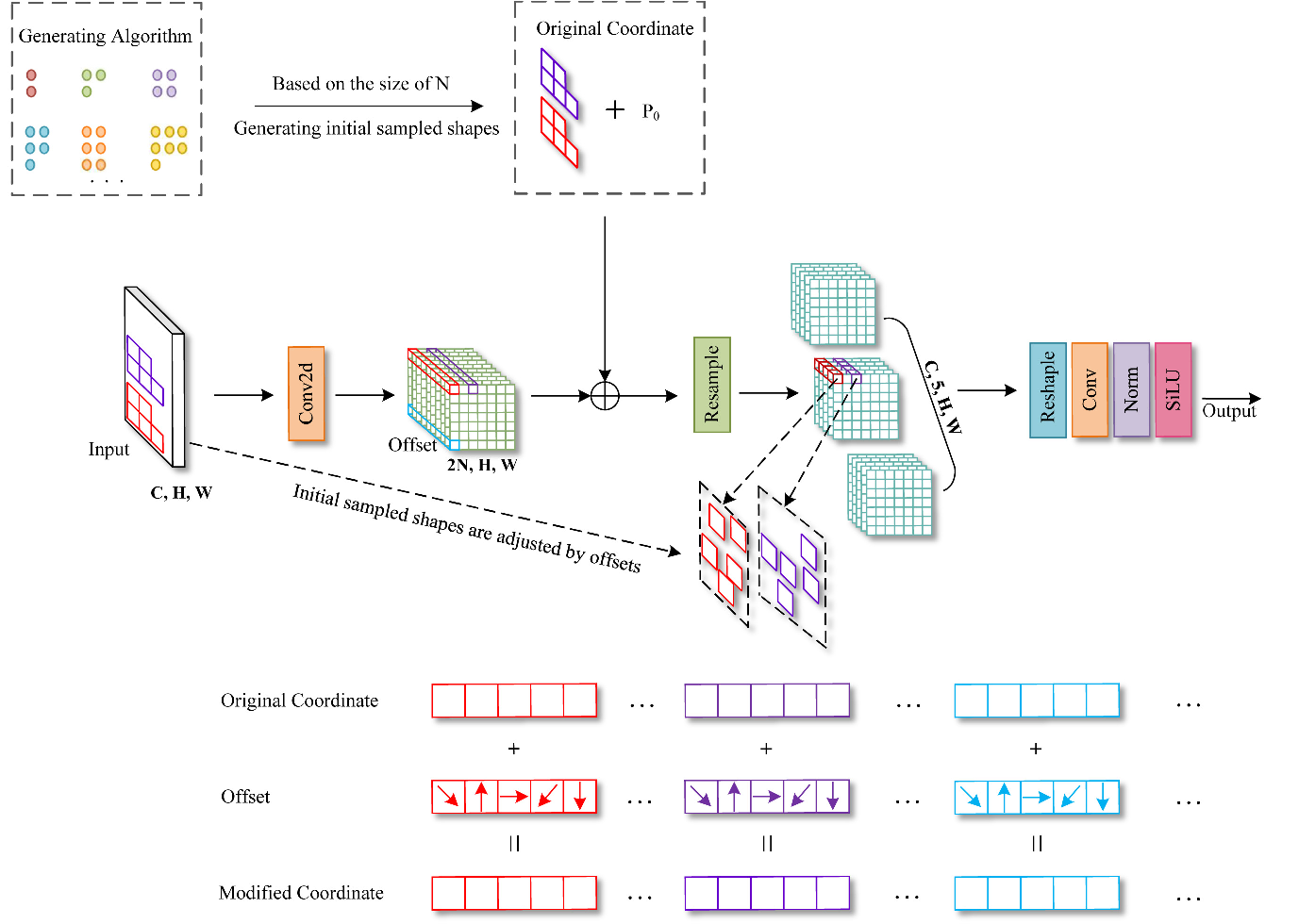}
	\caption{It shows a detailed schematic of the structure of LDConv. It assigns initial sampled coordinates to a convolution of arbitrary size and adjusts the sample shape with the learnable offsets. Compared to the original sampled shape, the sampled shape at each position is changed by resampling.}
	\label{LDConv}
\end{figure*}
$R$,  $w$,  $x\left(P_o+P_n\right)$ denotes the generated sampling grid, the convolutional parameter and the pixel at the corresponding position of the value. However, the irregular convolution operations are impossible to realize, because irregular sampled coordinates cannot be matched to the corresponding size convolution operations, e.g., convolution of sizes 5, 7, and 13. Cleverly, the proposed LDConv realizes it.

\begin{figure*}[h]
	\centering
	\includegraphics[trim=0 0 0 0,clip,scale=0.74]{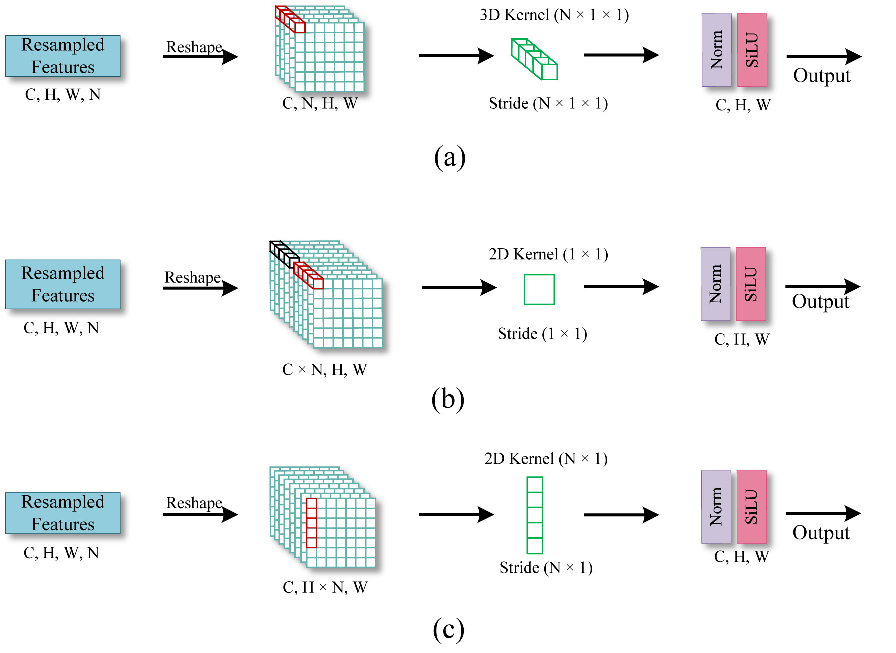}
	\caption{Three methods to extract features corresponding to the irregular convolutional kernels. The features boxed in red and black represent the convolutional kernel sampling features on the different channels.	(a) represents the Conv3d to complete it. (b) displays the 1 $\times$ 1 Conv2d to finish it. (c) shows the column Conv2d to solve it. }
	\label{get_feature}
\end{figure*}

\subsection{Linear deformable convoluiton}
It is obvious that the standard convolutional sampled position is fixed, which leads to the convolution can only extract the local information, and can not capture information of other positions. Deformable Conv learns the offsets through convolutional operations to adjust the sampled grid of the initial regular pattern. The approach compensates for the shortcomings of the convolution operation to a certain extent. However, the standard convolution and Deformable Conv are regular sampled grids that do not allow convolutional kernels with an arbitrary number of parameters. Moreover,  with an increase in the size of the convolutional kernel, the number of parameters tends to increase by a square, which is not friendly for the hardware environment. Therefore, we propose a novel Linear Deformable Convoluiton (LDConv). As shown in Figure~\ref{LDConv}, it illustrates the overall structure of The LDConv. The process of feature extraction by LDConv can be divided into three steps.

Take N = 5 as an example in Figure~\ref{LDConv}. Firstly, based on the size of N, the initial shaped shapes ($ P_{n}$) are generated by the proposed Algorithm~\ref{alg:code}. Then, the original coordinates ($P_{o} + P_{n}$) are obtained. This step focuses on generating the corresponding sampling coordinates on the feature map for a convolution kernel with N parameters. Secondly, offsets of the corresponding kernel are obtained by convolution operations, which have the dimensions (B, 2N, H, W), and then offsets are added to the original coordinates to generate the new sampling coordinates corresponding to the convolution. This step focuses on generating different sample shapes for the convolution at different locations in the feature map. Finally, the features at the corresponding positions are obtained by interpolating and resampling. Then apply the corresponding convolution operation to extract the features. This step is mainly to extract the features at the corresponding locations. Through the above three steps, LDConv can complete any size convolution operation to extract features.

The common point between LDConv and Deformable Conv is that both convolutions adjust the initial sampling shape by offsets. They differ in that LDConv completes the process of extracting features with an arbitrary number of parameters by generating initial sampling coordinates with the proposed algorithm. This is important because it corrects the growth of the number of Deformable Conv parameters to a linear growth. This provides flexibility in regulating the number of parameters and computational overhead of the network, as well as the consumption of device memory. Moreover, the last step of the convolution operation of Deformable Conv is unable to extract the features corresponding to irregular convolution. 

Instead, we explored three approaches in designing LDConv, as shown in Figure~\ref{get_feature}.
 Resampled features can be transformed into four dimensions (C, N, H, W), and then use Conv3d with step size and convolution size is (N, 1, 1) to extract the features. It can be represented in Figure~\ref{get_feature} (a). Resampled features also stack on the channel dimension (C $\times$ N, H, W), and then use the 1$ \times $1 convolution operation to reduce the dimension to (C, H, W).  It can be represented in Figure~\ref{get_feature} (b). In Deformable Conv \cite{dai2017deformable} and RFAConv \cite{zhang2023rfaconv} , they stack the 3 $\times$ 3 convolutional features in spatial dimensions. Then, a convolution operation with a step size of 3 is used to extract the features. However, this method targets square-sampled shapes, i.e., the convolution kernels are squares, such as 1 $\times$ 1, 2 $\times$ 2, 3 $\times$ 3, etc. Therefore, these methods cannot achieve irregular convolution operations. While, by stacking resampled features on rows or columns, features corresponding to irregular sample shapes can be extracted using column convolution or row convolution. The features are extracted to using a convolutional kernel of the appropriate size and step size. It is shown in Figure~\ref{get_feature} (c). These methods mentioned above can extract features corresponding to irregularly sampled shapes. It is only necessary to reshape features and use the corresponding convolution operation. Therefore, in Figure~\ref{LDConv}, the final "Reshape" and "Conv" represent any of the above methods. Moreover, to clearly show the process of LDConv, after resampling in Figure~\ref{LDConv}, we put the dimension of the feature corresponding to the convolutional size in the third dimension, however, when the code is implemented, it is located in the last dimension.

LDConv can perfectly accomplish the irregular convolutional feature extraction process, and it can flexibly adjust the sampled shape according to offsets and bring more exploration options for convolutional sampled shapes. It is more versatile as opposed to Standard Convolution and Deformable Conv, which are limited by the idea of a regular convolution kernel.

\begin{figure}[h]
	\centering
	\includegraphics[trim=0 0 0 0,clip,scale=1.3]{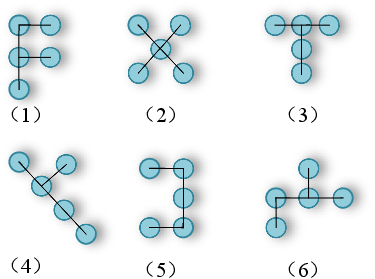}
	\caption{It shows the initial sampled shape of size 5.  LDConv can achieve arbitrary sampled shapes by designing different initial sampled shapes.}
	\label{LDConv5}
\end{figure}
\begin{figure}[h]
	\centering
	\includegraphics[trim=0 0 0 0,clip,scale=0.75]{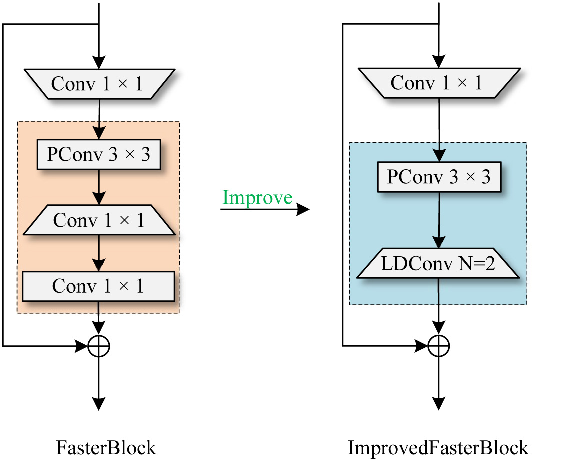}
	\caption{It shows the ImprovedFasterBlock. The LDConv (N = 2) is used to replace two 1 $\times$ 1 convolutional operations in FasterBlock.}
	\label{ImprovedFasterBlock}
\end{figure}
\begin{figure*}[h]
	\centering
	\includegraphics[trim=0 0 0 0,clip,scale=0.9]{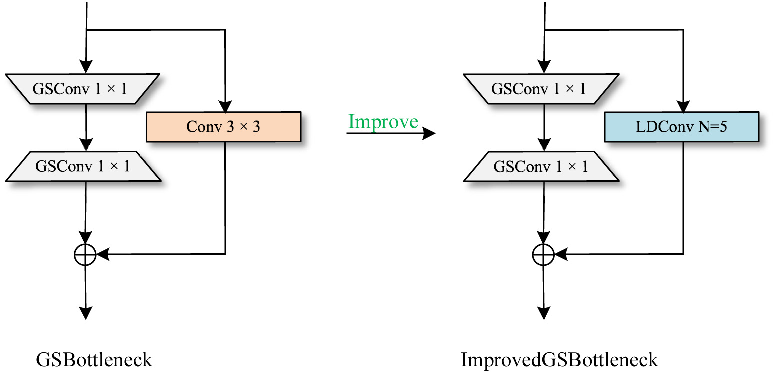}
	\caption{The ImprovedGSBottleneck is created by using the LDConv (N = 5) to replace the 3 $\times$ 3 convolutional operation and improve the GSBottleneck. }
	\label{ImprovedGSConvBottleneck}
\end{figure*}

\subsection{Extended LDConv}
We consider the design of LDConv to be innovative to accomplish the feat of extracting features with convolutional kernals of irregular and arbitrary sampled shapes. Even without using the offset idea in Deformable Conv, LDConv can still make a variety of convolution kernel shapes. Because, LDConv can resample the initial coordinates to present a variety of changes, as shown in Figure~\ref{LDConv5}, we design various initial sampled shapes for convolution of size 5. In Figure~\ref{LDConv5}, we only show some examples of size 5. However, the size of LDConv can be arbitrary, therefore as the size increases, the initial convolutional sampled shapes of LDConv become richer and even infinite. Given that the target shape varies across datasets, it is crucial to design the convolution operation corresponding to the sampled shape. LDConv fully realizes it by designing the convolution operation with the corresponding shape according to the phase-specific domain. It can also be similar to Deformable Conv by adding a learnable offset to dynamically adapt to changes in the object. For a specific task, the design of the initial sampled location of the convolution kernel is important, because it is a prior knowledge. As in Qi et al. \cite{qi2023dynamic}, they proposed sampled coordinates with corresponding shapes for the elongated tubular structure segmentation task, but their shape selection was only for elongated tubular structures.

LDConv really achieves the process of convolution kernel operation with any number of arbitrary shapes, and it can make the convolution kernel present a variety of shapes. Deformable Conv \cite{dai2017deformable} was designed to compensate for the shortcomings of regular convolution, while DSConv \cite{qi2023dynamic} was designed for specific object shapes. They have not explored convolution of arbitrary size and convolution of arbitrary sampled shapes.
The design of LDConv remedies these problems by allowing the convolution operation to efficiently extract the features of irregular sampled shapes through Offset. LDConv allows the convolution to have any number of convolution parameters, and allows the convolution to take on a wide variety of shapes. 

\begin{table*}[!h]
	\centering
	\footnotesize
	\renewcommand\arraystretch{1.3}
	\caption{Indicators $AP_{50}$, $AP_{75}$, $AP$, $AP_{S}$, $AP_{M}$, and $AP_{L}$ obtained on COCO2017 validation sets by networks constructed from different sizes of LDConv.}
	\setlength{\tabcolsep}{2mm}{
		\begin{tabular}{lccccccccccc}
			\toprule
			Models& LDConv & $AP_{50}$(\%) & $AP_{75}$(\%) &  $AP(\%)$ & $AP_{S}$(\%) & $AP_{M}(\%)$ & $AP_{L}(\%)$ & GFLOPS & Params (M)& Time (ms)\\
			\midrule
			YOLOv5n (Baseline)	& -      & 45.6     & 28.9     & 27.5          & 13.5          & 31.5   & 35.9         & 4.5          & 1.87  &   4.4    \\
			\midrule
			\multirow{4}{*}{Improved-YOLOv5n}
			& 3      & 47.8     & 31.1     & 29.8          & 14.5          & 33.2   & 41.0         & 3.8          & 1.51      &      4.8\\
			& 5      & 48.8     & 32.6     & 31.0          & 14.6          & 34.1   & 43.2         & 4.1          & 1.65      &     4.7 \\
			& 9      & 50.5     & 33.9     & 32.3          & 14.9          & 36.1   & 44.1         & 4.8          & 1.94     &       5.1\\
			& 13     & 51.2     & 34.5     & 33.0          & 15.7          & 36.3   & 45.6         & 5.5          & 2.23      &      5.7\\
			\midrule
			
			YOLOv5s (Baseline)& -      & 57.0       & 39.9     & 37.1          & 20.9          & 42.4   & 47.8         & 16.4         & 7.23   &     4.8    \\
			\midrule
			\multirow{3}{*}{Improved-YOLOv5s}
			& 4      & 58.2     & 41.9     & 39.2          & 21.4          & 43.2   & 53.4         & 14.1         & 6.01      &  5.1    \\
			& 6      & 59.2     & 42.6     & 39.9          & 21.5          & 44.2   & 54.7         & 15.3         & 6.55       &    5.8 \\
			& 7      & 59.4     & 43.2     & 40.4          & 21.5          & 44.6   & 55.1         & 15.9         & 6.82        &  6.1  \\
			\bottomrule
	\end{tabular}}
	\label{detect1}
\end{table*}
Moreover, LDConv as a flexible convolution operation, it can easily replace the convolution operation in some efficient modules to improve network performance, such as  FasterBlock \cite{chen2023run}, and GSBlottleneck \cite{li2022slim}. Therefore, the ImprovedFasterBlock and ImprovedGSBlottleneck are proposed, as shown in  Figure~\ref{ImprovedFasterBlock} and Figure~\ref{ImprovedGSConvBottleneck}.  In FasterBlock, the PConv is firstly utilized to extract features, then two 1 $\times$ 1 convolutional operation to construct Bottleneck. Therefore, the LDConv (N = 2) is used to replace them and ensure the number of parameters. In GSBottleneck, the LDConv (N = 5) is used to replace the 3 $\times$ 3 convolutional operation to improve the performance and reduce the number of parameters and computational overhead.

\section{Experiments}

To verify the advantages of LDConv, we conduct rich target detection experiments based on advanced YOLOv5 \cite{YOLOV5}, YOLOv7 \cite{wang2023yolov7} and YOLOv8 \cite{YOLOV8} respectively. Firstly, YOLO series algorithms, as representative algorithms in the field of target detection, have been widely used in various fields and have obtained good detection performance. Secondly, the network structure of the YOLO family of algorithms is simple compared to other network structures, which facilitates us to do a large number of experiments on different datasets to demonstrate the advantages of LDConv. All models in the experiments are trained based on RTX3090. To validate the advantages of LDConv we perform related experiments on representative COCO2017, VOC 7 + 12 and VisDrone-DET2021 datasets respectively.

\begin{table*}[h!]
	\centering
	\footnotesize
	\renewcommand\arraystretch{1.3}
	\caption{Experimental results obtained with networks constructed from LDConv with different size based on VOC 7+12. }
	\setlength{\tabcolsep}{2mm}{
		\begin{tabular}{lcccccccc}
			\toprule
			Models & LDConv     & Precision(\%)
			& Recall(\%) & mAP50(\%)    & mAP(\%) & GFLOPS & Params (M)  &Time (ms)              \\
			\midrule
			YOLOv7-tiny (Baseline)& -          & 77.3         & 69.8    & 76.4   & 50.2      & 13.2 & 6.06 & 2.5 \\
			\midrule
			\multirow{6}{*}{Improved-YOLOv7-tiny}
			& 3          & 80.1         & 68.4    & 76.1   & 50.3      & 12.1 & 5.56  & 2.0 \\
			& 4          & 78.2         & 70.3    & 76.2   & 50.7      & 12.4 & 5.66  & 2.1 \\
			& 5          & 77.0         & 71.1    & 76.5   & 50.8      & 12.6 & 5.75  & 2.3 \\
			& 6          & 79.6         & 69.9    & 76.9   & 51.0      & 12.9 & 5.85  & 2.5 \\
			& 8          & 78.6         & 70.1    & 76.7   & 51.2      & 13.4 & 6.04  & 3.1\\
			& 9          & 81.0         & 69.3    & 76.7   & 51.3      & 13.7 & 6.14  & 3.2\\
			\bottomrule
	\end{tabular}}
	\label{detect2}
\end{table*}

\subsection{Object detection experiments on COCO2017}
COCO2017 includes train set (118287 images), val set (5000 images), and covers 80 object classes. It has become a standard dataset in the field of computer vision research, especially in the field of target detection. We choose the state-of-the-art YOLOv5n and YOLOv5s detectors as the baseline model. Then, LDConv with different sizes is used to replace the convolution operations of YOLOv5n and YOLOv5s. The replacement details are the same as the target detection experiments in \cite{zhang2023rfaconv}. In the experiments, the default parameters of the network are used except for the epoch and batch-size parameters. Based on a batch size of 32, we trained each model for 300 epochs. Following previous work, we report $AP_{50}$, $AP_{75}$, AP, $AP_{S}$, $AP_{M}$ and  $AP_{L}$.

 Moreover, we also report target detection on YOLOv5n and YOLOv5s for LDConv with sizes 5, 4, 6, 7, 9, and 13, respectively. As shown in Table~\ref{detect1}, the detection accuracy of YOLOv5 gradually increases with the increase of the convolutional kernel size, while the number of parameters required by the model and the computational overhead also gradually increase. Time in the Table~\ref{detect1} indicates the time required to process an image, expressed in milliseconds (ms). Compared to standard convolutional operations, LDConv substantially improves the target detection performance of YOLOv5 on COCO2017. It can be seen that when the size of LDConv is 5, it not only makes the number of parameters and computational overhead required by the model decrease, but also significantly improves the detection accuracy of YOLOv5n. Its $AP_{50}$, $AP_{75}$, and also AP are all improved by three percentage points, which is outstanding. LDConv improves the $AP_{S}$, $AP_{M}$, and $AP_{L}$ of the baseline model, but it is obvious that LDConv improves the detection accuracy of large objects significantly compared to small and middle objects. We assert that LDConv uses offsets to better adapt to the shape of large objects.
\begin{table*}[h]
	\centering
	\footnotesize
	\caption{Indicators Precision, Recall, mAP50 and mAP obtained on VisDrone-DET2021 validation sets by networks constructed from different sizes of LDConv.}
	\renewcommand\arraystretch{1.3}
	\setlength{\tabcolsep}{2mm}{
		\begin{tabular}{lcccccccc}
			\toprule
			Models & LDConv     & Precision(\%) & Recall(\%) & mAP50(\%)    & mAP(\%) & GFLOPS & Params (M)&Time (ms)              \\
			\midrule
			YOLOv5n (Baseline)   &   -          & 38.5         & 28      & 26.4   & 13.4      & 4.2   & 1.77&9.5 \\
			\midrule
			\multirow{7}{*}{Improved-YOLOv5n}
			& 3          & 37.9         & 27.4    & 25.9   & 13.2      & 3.5   & 1.41 &12.1\\
			& 5          & 40.0         & 28.0    & 26.9   & 13.7      & 3.8   & 1.56 &12.9\\
			& 6          & 38.1         & 28.1    & 26.8   & 13.6      & 4.0   & 1.63 &10.8\\
			& 7          & 39.8         & 28.2    & 27.5   & 14.2      & 4.2   & 1.70 &10.6\\
			& 9          & 39.7         & 28.9    & 27.7   & 14.3      & 4.5   & 1.84 &11.1\\
			& 11         & 40.4         & 28.8    & 27.7   & 14.2      & 4.8   & 1.99 &10.8\\
			& 14         & 40.0         & 28.8    & 27.9   & 14.3      & 5.3   & 2.20 &10.9\\
			\bottomrule
	\end{tabular}}
	\label{detect3}
\end{table*}

\subsection{Object detection experiments on VOC 7+12}

In order to further validate our method, we conduct experiments on the VOC 7+12 dataset, which is a combination of VOC2007 and VOC2012, comprising 16551 training images and 4952 validating images, and covers 20 object categories. To test the generalizability of LDConv across different architectures, we selected YOLOv7-tiny as the baseline model. Since YOLOv7 and YOLOv5 are systems with different architectures, it is possible to compare the performance of LDConv with different architectural settings. In YOLOv7-tiny, we use LDConv with different sizes to replace standard convolutional operation. The details of the replacement follows the work in \cite{zhang2023rfaconv}. The hyperparameter settings for all models are consistent with those in the previous section. Following previous work, we present both mAP50 and mAP. As demonstrated in Table~\ref{detect2}, with the increasement of size in LDConv, the network's detection accuracy gradually improves, while the model's parameter count and computational demand also incrementally rise. These experiments further substantiate the advantages of LDConv.

\subsection{Object detection experiments on VisDrone-DET2021}
In order to verify again that LDConv has strong generalization ability, based on VisDrone-DET2021 data, we conducted relevant target detection experiments. VisDrone-DET2021 is a challenging dataset taken by UAVs in different environments, weather and lighting conditions. It is one of the largest datasets with the widest coverage of UAV aerial photography in China. The number of images in the training sets is 6471 and the number of images in validation sets is 548. As in Section 4.1, we choose YOLOv5n as the baseline to use LDConv to replace convolutional operations in the network. In experiments, the batch-size is set to 16 to facilitate the exploration of larger convolution sizes, and all other hyperparameter settings are the same as before. As in the previous section, we report mAP50 and mAP, respectively. As shown in Table~\ref{detect3}, it is clear to see that LDConv based on different sizes can be used as a lightweight option to reduce the number of parameters and computational overhead and improve network performance. In experiments, when the size of LDConv is set to 3, the detection performance of the model decreases compared to the baseline model, but the corresponding number of parameters and computational overhead are much smaller. Moreover, we can gradually adjust the size of LDConv to explore the changes in network performance. LDConv brings richer options to the network.

\begin{table*}[!h]
	\centering
	\footnotesize
	\caption{Indicators $AP_{50}$, $AP_{75}$, $AP$, $AP_{S}$, $AP_{M}$, and $AP_{L}$ obtained on COCO2017 validation sets by networks constructed from different sizes of LDConv, DSConv and Deformable Conv.}
	\renewcommand\arraystretch{1.3}
	\setlength{\tabcolsep}{1.55mm}{
		\begin{tabular}{lccccccccc}
			\toprule
			Models                 & $AP_{50}(\%)$ & $AP_{75}(\%)$ & $AP(\%)$   & $AP_{S}(\%)$  & $AP_{M}(\%)$  & $AP_{L}(\%)$  & GFLOPS & Params (M) & Time (ms) \\
			\midrule 
			YOLOv5s                                &54.8    &37.5    &35.0    &19.2   &40.0     &45.2   &16.4     &7.23    &4.0 \\
			YOLOv5s (DSConv=5)                     &43.2    &23.5    &23.9  &13.0   &27.6   &30.5   &14.8     &6.45    &7.7 \\
			YOLOv5s (LDConv=5)             &56.6    &40.7    &38.0  &20.8   &41.8   &52.0   &14.8     &6.45    &6.7 \\
			YOLOv5s (LDConv=9)             &57.8    &41.4    &38.7  &20.8   &42.8   &52.3   &17.1     &7.37    &7.7 \\
			YOLOv5s (LDConv=9, Padding=1)  &58.3    &41.9    &39.2  &21.6   &43.2   &53.5   &17.1     &7.37    &6.9 \\
			YOLOv5s (Deformable Conv=3)            &58.5    &41.8    &39.1  &20.8   &43.4   &53.6   &17.1     &7.37    &7.4 \\
			YOLOv5s (LDConv=11)            &58.5    &42.1    &39.3  &21.9   &43.3   &53.8   &18.3     &7.91    &7.7 \\
			YOLOv5s (LDConv=11, Padding=1) &58.6    &42.1    &39.5  &21.3   &43.7   &53.2   &18.3     &7.91    &7.4 \\	
			\bottomrule
	\end{tabular}}
	\label{detect4}
\end{table*}

\begin{table*}[h]
	\centering
	\footnotesize
	\caption{Indicators Precision, Recall, mAP50 and mAP obtained on VOC 7+12 validation sets by networks constructed from different sizes of LDConv, DSConv and Deformable Conv.}
	\renewcommand\arraystretch{1.3}
	\setlength{\tabcolsep}{2mm}{
	\begin{tabular}{lccccccc}
			\toprule
			Models             & Precision(\%) & Recall(\%) & mAP50(\%) & mAP(\%) & GFLOPS & Params(M) & Time (ms) \\
			\midrule
			YOLOv5n						& 73.8			&62.2       & 68.1      & 41.5    & 4.2    & 1.78   &2.8\\
			YOLOv5n (DSConv=4)			& 63.0			&50.4       & 54.2      & 26.1    & 3.7    & 1.55   &2.7\\
			YOLOv5n (LDConv=4)	        & 76.5			&63.6       & 70.8      & 46.5    & 3.7    & 1.55   &2.7\\
			YOLOv5n (DSConv=9)			& 60.6			&50.8       & 53.4      & 25.3    & 4.8    & 1.90   &4.6\\
			YOLOv5n (Deformable Conv=3)	& 77.3			&64.0       & 71.6      & 48.0    & 4.8    & 1.90   &4.3\\
 			YOLOv5n (LDConv=9)         	& 76.7			&65.2       & 71.8      & 48.4    & 4.8    & 1.90   &4.2\\
			\bottomrule
	\end{tabular}}
	\label{detect5}
\end{table*}

\begin{table*}[h]
	\centering
	\footnotesize
	\caption{Indicators Precision, Recall, mAP50 and mAP obtained on VOC 7+12 validation sets by networks constructed from different convolutional methods.}
	\renewcommand\arraystretch{1.3}
	\setlength{\tabcolsep}{2.5mm}{
		\begin{tabular}{lccccccc}
			\toprule
			Models                   & Precision(\%) & Recall(\%) & mAP50(\%) & mAP(\%)  & GFLOPS & Params (M) &Time (ms) \\
			\midrule
			BaseLine \cite{YOLOV5} (2020)          &73.8       & 62.2    & 68.1 & 41.5  & 4.2     & 1.77 &2.8  \\
			SAConv \cite{qiao2021detectors} (2021) &76.7       & 64.2    & 70.8 & 45.2  & -       & 2.41 &2.8    \\
			GSConv-Neck \cite{li2022slim} (2022)   &74.3       & 61.4    & 68.0   & 42.3 & 4.0    & 1.68 &2.7   \\
			SPDConv  \cite{sunkara2022no} (2022)   &75.1       & 64.1    & 70.6 & 44.8  & 7.8     & 3.51 &2.7     \\
			CKConv \cite{DBLP:conf/iclr/RomeroKBTH22}(2022)    &73.0     &60.5  &66.2    &39.0    &16.2  &3.20 &1.8\\
			FlexConv \cite{flexConv}(2022)         &53.8       &46.8     &47.3    &21.1    &4.4   &3.20 &2.4\\
			RFAConv \cite{zhang2023rfaconv} (2023) &75.0       & 62.9    & 69.3 & 43.1  & 4.4     & 1.82 &2.9   \\
			FasterBlock \cite{chen2023run} (2023)  &71.7       & 60.2    & 65.9  & 39.3    & 3.6  & 1.54 &2.7 \\
	    	DCLS \cite{hassani2023dilated} (2023)&71.8       & 59.0    & 65.1  & 38.0    & 2.9  & 1.79 &4.0 \\
			LDConv (Ours)                  &76.5       & 63.6    & 70.8 & 46.5   & 3.7   & 1.55  &2.7\\
			\midrule
			ImprovedGSConv-Neck                    &75.2       & 63.2    & 69.5 & 46.0  & 3.8  & 1.53  &4.0  \\
			ImprovedFasterBLock                    &76.3       & 62.5    & 69.8 & 45.7  & 3.4   & 1.45 &3.4  \\
			\bottomrule
	\end{tabular}}
	\label{detect8}
\end{table*}

\begin{table*}[h]
	\centering
	\footnotesize
	\caption{Indicators Precision, Recall, mAP50 and mAP obtained on VisDrone-DET2021 validation sets by networks constructed from different convolutional methods.}
	\renewcommand\arraystretch{1.3}
	\setlength{\tabcolsep}{2.5mm}{
		\begin{tabular}{lccccccc}
			\toprule
			Models                                 & Precision(\%) & Recall(\%) & mAP50(\%)   & mAP(\%)   & GFLOPS       & Params (M)    &Time (ms)\\
			\midrule
			BaseLine \cite{YOLOV5}(2020)           &45.1      & 33.5     & 32.9   & 17.9       &15.8    & 7.0     &9.0 \\
			SAConv \cite{qiao2021detectors} (2022) &45.0       & 33.1    & 32.9   & 17.7       & -      & 9.53    &9.2 \\
			GSConv-Neck \cite{li2022slim} (2022)   &43.7      & 31.7     & 31.6   & 16.7       & 14.6   & 6.53    &9.2\\
			SPDConv \cite{sunkara2022no} (2022)    &44.3      & 34.0     & 33.3   &18.0        & 30.0   & 13.95   &9.4 \\
			CKConv \cite{DBLP:conf/iclr/RomeroKBTH22}(2022)    &43.9     &31.9    &31.5        &16.6    &63.1  &12.5 &10.9\\
			FlexConv \cite{flexConv}(2022)         &37.6       &30.3     &27.6    &13.3        &16.5   &12.5     &13.4\\
			RFAConv  \cite{zhang2023rfaconv} (2023)&45.6      & 34.1     & 33.9   &18.3        & 16.3   & 7.13    &9.3    \\
			FasterBlock \cite{chen2023run} (2023)  &44.7      & 33.6     & 33.2   & 17.9       & 13.3   & 6.03    &9.2 \\
			DCLS \cite{hassani2023dilated} (2023)  &45.5       & 32.8    & 32.6  & 17.2        &10.7  & 7.04      &17.9 \\
			LDConv (Ours)                  &45.2      & 34.0     & 33.4   & 17.9       & 14.1   & 6.26    &10.8 \\
			\midrule
			ImprovedGSConv-Neck                    &45.1      & 32.7     &32.5    & 17.2       & 12.9   & 5.75    &10.8 \\
			ImprovedFasterBLock                    &46.8      & 33.2     &33.8    & 18.0       & 12.4   & 5.63    &9.6 \\
			
			\bottomrule
	\end{tabular}}
	\label{detect9}
\end{table*}

\begin{figure*}[!h]
	\centering
	\begin{subfigure}
		\centering
		\includegraphics[width=0.49\linewidth]{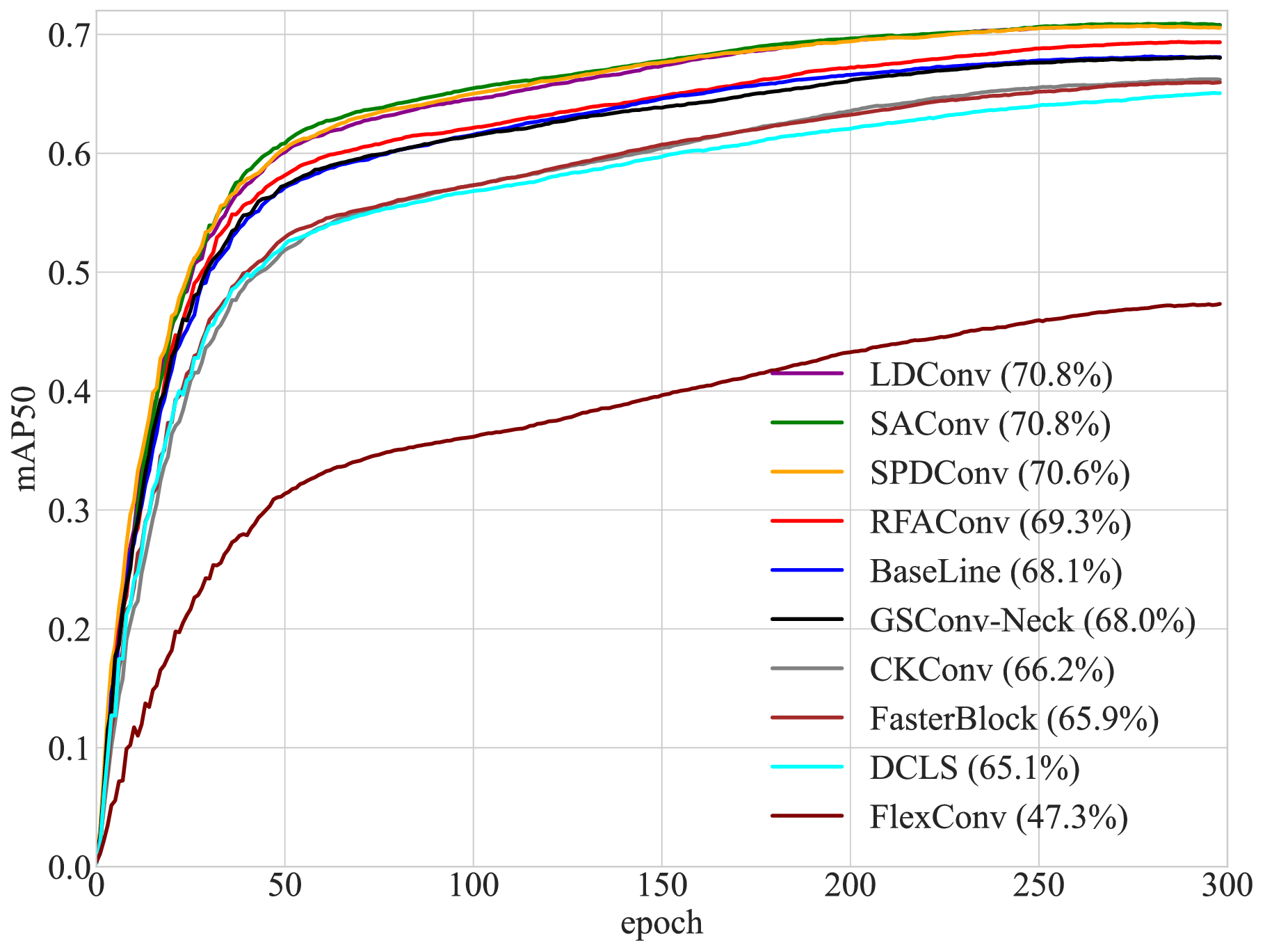}
	\end{subfigure}
	\centering
	\begin{subfigure}
		\centering
		\includegraphics[width=0.49\linewidth]{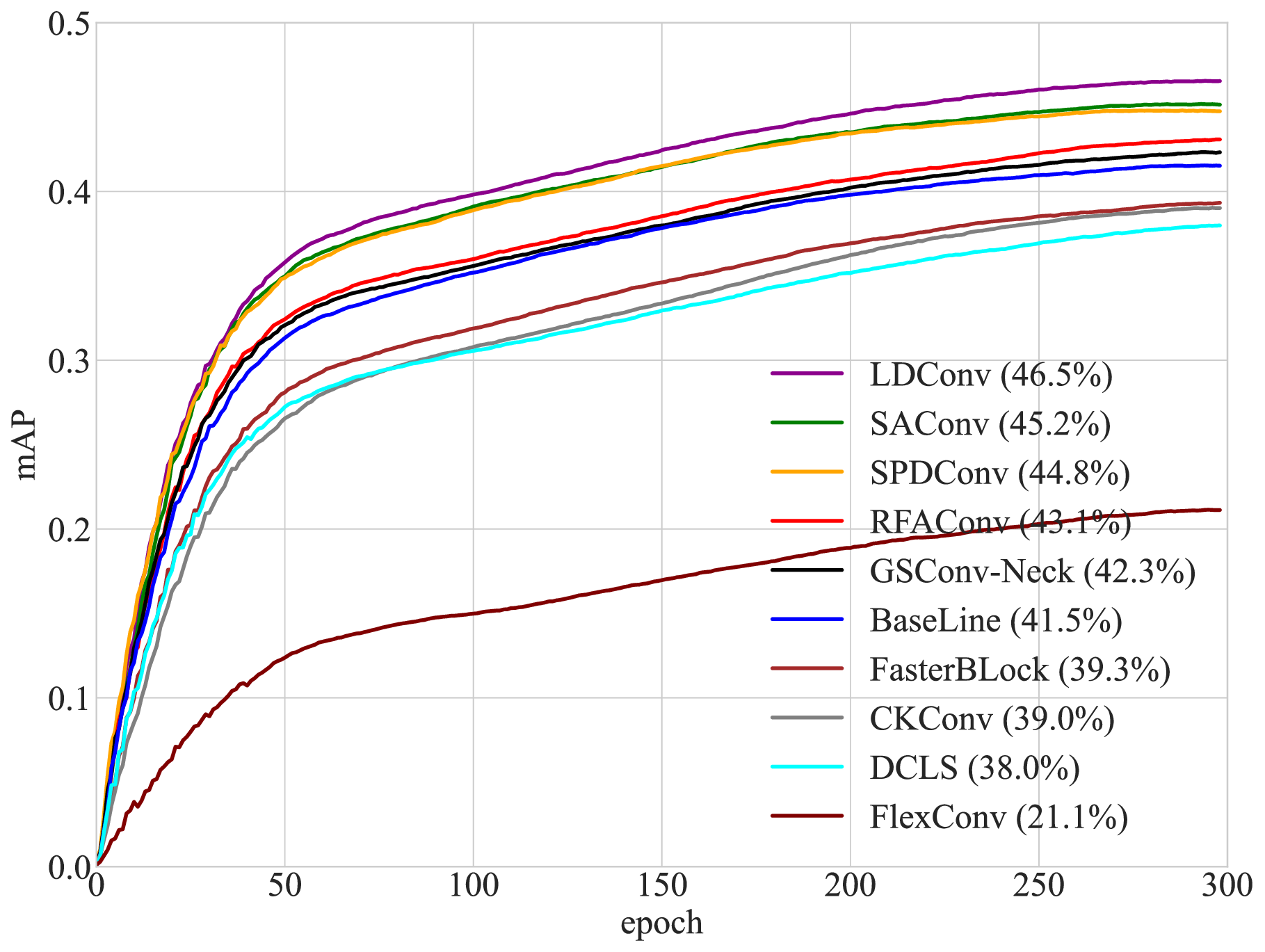}
		\caption{Based on VOC 7+12, changes in mAP50 and mAP during training for different methods. }
		\label{map-VOC}
	\end{subfigure}
\end{figure*}
\begin{figure*}[h]
	\centering
	\begin{subfigure}
		\centering
		\includegraphics[width=0.49\linewidth]{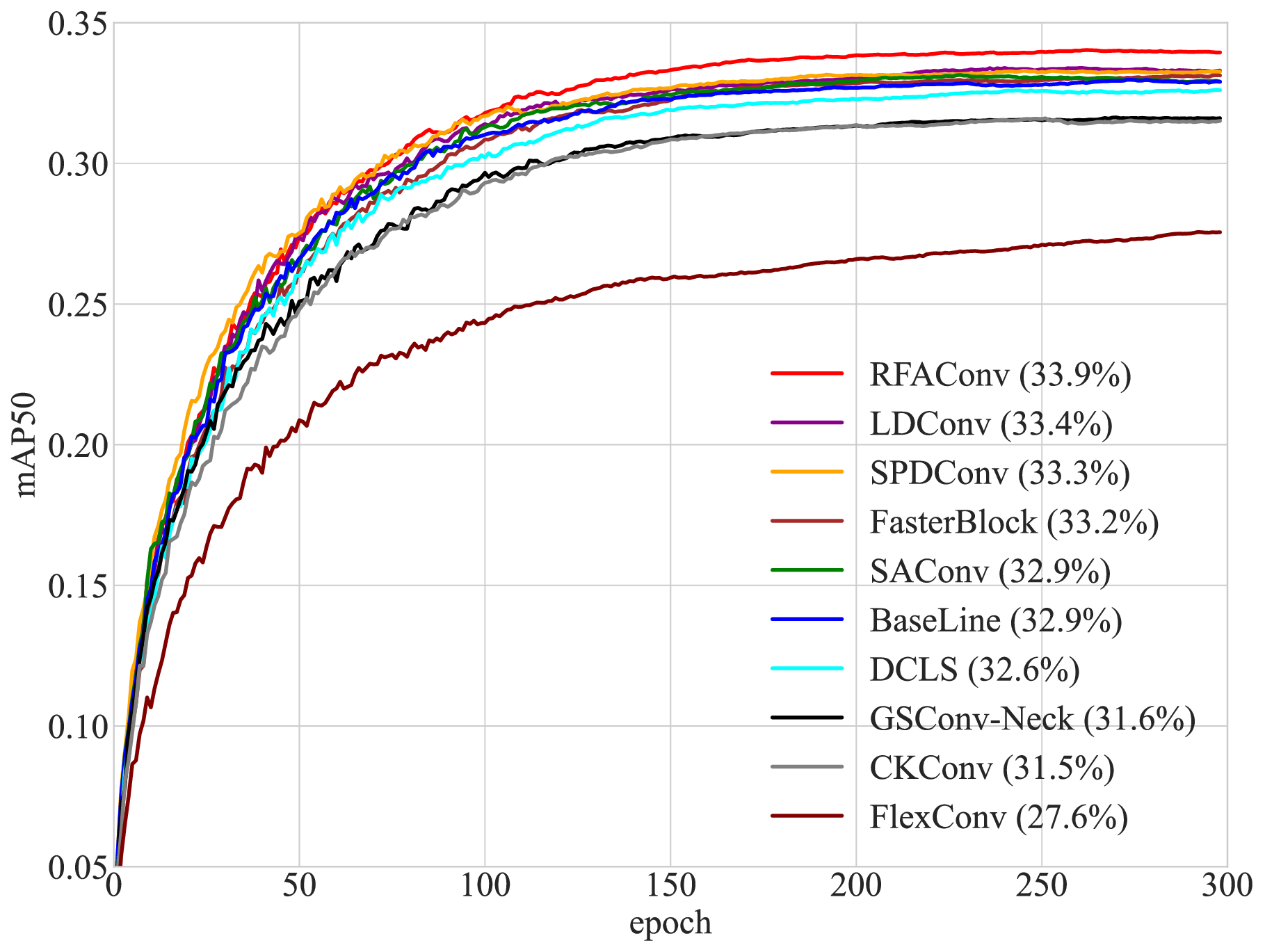}
	\end{subfigure}
	\centering
	\begin{subfigure}
		\centering
		\includegraphics[width=0.49\linewidth]{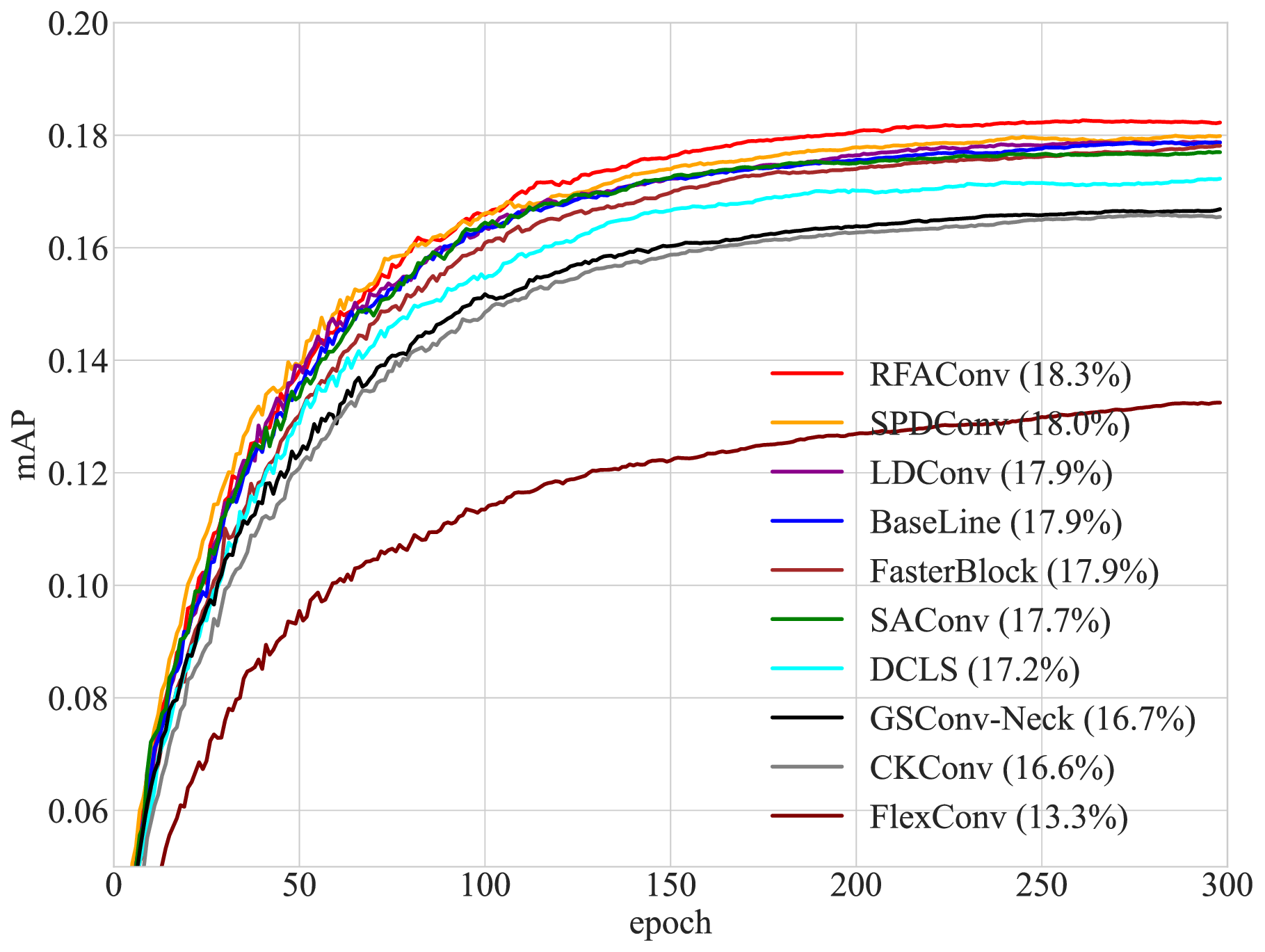}
		\caption{Based on VisDrone-DET2021, changes in mAP50 and mAP during training for different methods.}
		\label{map-VisDrone}
	\end{subfigure}
\end{figure*}

\subsection{Comparison experiments}
Unlike Deformable Conv \cite{dai2017deformable}, LDConv offers a richer choice for networks and compensates for the shortcomings of Deformable Conv, which only uses regular convolution operations, while LDConv can use both regular and irregular convolution operations. When the size of LDConv is set to the square of K, it becomes a Deformable Conv. Moreover, DSConv \cite{qi2023dynamic} also uses offsets to adjust the sampled shapes, but its sampled shape is designed for tubular targets, and the change of the sampled shape is limited. To contrast the advantages of LDConv, Deformable Conv, and DSConv at the same size, we perform experiments in COCO2017 based on YOLOv5s. As shown in the Table~\ref{detect4}, when the number of convolution kernel parameters is 9 (That is the standard 3 $\times$ 3 convolution), it can be seen that the performance of LDConv and Deformable Conv is the same. Because when the convolution kernel size is regular, the LDConv is equivalent to the Deformable Conv. But we have mentioned that Deformable Conv has not explored the irregular convolution kernel size. Therefore, a convolution operation with a number of parameters of 5 or 11 cannot be implemented. In the design of LDConv, we do not implement zero-padding for input features. However, padding is used in Deformable Conv. For a fair comparison, we also utilize zero-padding for input features in LDConv. Experiments show that zero-padding in LDConv helps the network to improve performance. 

Moreover, based on VOC 7+12, we chose the YOLOv5n to compare LDConv, DSConv, and Deformable Conv. In using Deformable Conv, the zero padding is eliminated to make a fair comparison with LDConv. As shown in Table~\ref{detect5}, it can be seen that LDConv achieves better performance. Theoretically, the performance of Deformable Conv should be equivalent to LDConv (N = 9) when the size of Deformable Conv is 3. However, Deformable Conv samples from index = 1 instead of 0, therefore, some information is lost. Since DSConv is designed for a specific tubular shape, it can be seen that its detection performance on COCO2017 and VOC 7 + 12 is not obvious. 

When implementing DSConv, Qi et al. \cite{qi2023dynamic} expand the features of rows or columns, and finally use row or column convolution to extract features similar to us. So their method can also implement convolution operations with parameters 2, 3, 4, 5, 6, 7, etc. Under the same size, we also conduct a comparison experiment. Because, the DSConv does not completes the down-sample method,  in experiments, we use the LDConv and DSConv to replace 3 $\times$ 3 convolution of C3 in YOLOv5n. Experimental results are shown in Table~\ref{detect4} and Table~\ref{detect5}. LDConv is advantageous over DSConv, because DSConv is not designed to improve the performance of convolutional kernels of arbitrary size, but rather to explore for targets of specific shapes. In contrast, LDConv provides a richer choice of convolutional kernel selection and exploration that can effectively improve network performance. 

and Table~\ref{detect5}

 Moreover, to highlight the advantages of the LDConv, novel convolutional methods are compared, such as SAConv \cite{qiao2021detectors}, SPDConv \cite{sunkara2022no}, RFAConv \cite{zhang2023rfaconv}, FasterBlock \cite{chen2023run}, GSConv-Neck \cite{li2022slim} and related works CKConv \cite{DBLP:conf/iclr/RomeroKBTH22}, FlexConv \cite{flexConv} and DCLS \cite{hassani2023dilated}. Based on YOLOv5n and YOLOv5s, we conduct experiments on representative datasets VOC 7+12 and VisDrone-DET2021, respectively. Meanwhile, we implement experiments with ImprovedFasterBLock and ImprovedGSConv-Neck constructed by ImprovedGSBottleneck. All experiments are conducted based on a batch-size of 16 and an epoch of 300. In the experiments, the 3 $\times$ 3 convolutional operation in Bottleneck is replaced by LDConv, SAConv, FasterBlock, CKConv, FlexConv and DCLS, respectively. Following works \cite{zhang2023rfaconv,sunkara2022no,li2022slim}, the down-sampling convolution is replaced by RFAConv and SPDConv, and the C3 module is replaced by GSConv-Neck.

 As shown in Table~\ref{detect8} and Table~\ref{detect9}, it can be seen that the network using LDConv gets good performance. In Table~\ref{detect9}, Compared to LDConv, RFAConv achieves better performance on VisDrone-DET2021, which  is a challenging dataset that contains images acquired by drones. However, images captured by UAVs usually contain complex backgrounds. RFAConv assigns different attention weights to the convolution kernel at each sampling position, allowing the convolution to better distinguish background information when features from UAV-captured images. In contrast, LDConv does not have flexible attentional weights to distinguish background information. Moreover, CKConv and FlexConv are used as a flexible convolutional operation, and they can similarly be set with an arbitrarily sized memory range, but the implementation of the basis requires an excessive amount of computation and the networks obtain poor performance. We set the hyperparameter horizon of CKConv to 1, 3, and 5 in our experiments. The performance of the network is poor when the horizon is 1, and when the horizon is set to 3 the network obtains good performance compared to the horizon of 1, but at the expense of a large amount of computation. When horizon is 5, the network does not get very good performance but the computational overhead increases dramatically. Therefore, in our experiments, we set the horizon of CKConv to 3. FlexConv is based on CKConv, so to maintain uniformity, we set the horizon to 3 for the experiments as well.The DCLS replaces the standard 3 $\times$ 3 convolutional operation by referring to the corresponding repository settings.

In order to clearly observe the changes in mAP50 and mAP of all networks during the training process, we visualize their changes based on VOC 7+12 and VisDrone-DET2021. As shown in Figure~\ref{map-VOC} and Figure~\ref{map-VisDrone}. Compared to these state-of-the-art methods, LDConv is able to flexibly adjust the number of convolutional parameters and the sampled shapes to trade-off the network's overhead and adapt to changing targets, therefore good performance is obtained by adding LDConv to the network.

\begin{table*}[h]
	\centering
	\footnotesize
	\caption{Indicators $AP_{50}$, $AP_{75}$, $AP$, $AP_{S}$, $AP_{M}$, and $AP_{L}$ obtained on COCO2017 validation sets by networks constructed from different sampled shapes of LDConv. The "Sampled Shape i" denotes different initial sampled shapes of LDConv. }
	\renewcommand\arraystretch{1.3}
	\setlength{\tabcolsep}{2mm}{
		\begin{tabular}{lccccccccc}
			\toprule
			Models            & $AP_{50}(\%)$ & $AP_{75}(\%)$ & $AP(\%)$   & $AP_{S}(\%)$  & $AP_{M}(\%)$  & $AP_{L}(\%)$  & GFLOPS & Params (M) &Time (ms) \\
			\midrule
			YOLOv8n                     & 49.0 & 37.1 & 34.2 & 16.9 & 37.1 & 49.1 & 8.7    & 3.15  & 2.3    \\
			YOLOv8n-5 (Sampled Shape 1) & 49.5 & 37.6 & 34.9 & 16.8 & 38.2 & 50.2 & 8.4    & 2.94  & 3.1  \\
			YOLOv8n-5 (Sampled Shape 2) & 49.6 & 37.8 & 34.9 & 15.9 & 38.4 & 50.1 & 8.4    & 2.94  & 2.9  \\
			YOLOv8n-5 (Sampled Shape 3) & 49.6 & 38.1 & 35.0 & 16.6 & 38.2 & 50.9 & 8.4    & 2.94  & 2.9  \\
			YOLOv8n-6 (Sampled Shape 1) & 50.1 & 38.3 & 35.3 & 16.6 & 38.6 & 51.1 & 8.6    & 3.01  & 2.9  \\
			YOLOv8n-6 (Sampled Shape 2) & 50.2 & 38.2 & 35.4 & 16.6 & 38.3 & 51.3 & 8.6    & 3.01  & 3.0 \\
			
			\bottomrule
	\end{tabular}}
	\label{detect6}
\end{table*}

\begin{table}[h]
	\centering
	\footnotesize
	\caption{Experimental results obtained with networks constructed from LDConv with different sampling shapes based on VisDrone-DET2021.}
	\renewcommand\arraystretch{1.3}
	\setlength{\tabcolsep}{0.8mm}{
		\begin{tabular}{lccccc}
			\toprule
			Models       &Shapes      & Precision(\%) & Recall(\%) & mAP50(\%) & mAP(\%)  \\
			\midrule
			\multirow{5}{*}{YOLOv5n}
			&a & 39.5     & 27.9   & 26.9  & 13.7 \\
			&b & 39.4     & 28.2   & 26.8  & 13.6 \\
			&c & 37.4     & 27.8   & 26.1  & 13.4 \\
			&d & 37.5     & 27.0   & 25.5  & 12.9 \\
			&e & 38.4     & 27.6   & 26.4  & 13.4 \\
			\bottomrule
	\end{tabular}}
	\label{detect7}
\end{table}

\subsection{Exploring the initial sampled shapes}
As mentioned earlier, LDConv can extract features by using arbitrary sizes and arbitrary sampled shapes. To explore the effect of LDConv with different initial sampled shapes on the network, we conduct experiments at COC-O2017 and VisDrone-DET2021, respectively. On COCO2017, we conduct experiments based on a batch-size of 32 and an epoch of 100. In VisDrone-DET2021, we conducted experiments based on a batch-size of 16 and an epoch of 300. All other hyperparameters are network defaults. In COCO2017, we choose YOLOv8n for our experiments. As shown in Table~\ref{detect6}, LDConv can still improve the detection accuracy of the network. The network structures of YOLOv8 and YOLOv5 are similar. One of the main differences is the design of C3 and C2f. It can be seen that the increase in performance obtained by the addition of LDConv in YOLOv8 is not as good as that obtained in YOLOv5. We assert that YOLOv8 needs more parameters than YOLOv5 under the same size, so more number of parameters can provide better feature information as LDConv does. Therefore with the addition of LDConv, the YOLOv8 boost is not as significant as the YOLOv5. Furthermore, at the same size, we test the effect of different initial sampled shapes on network performance in COCO2017. It is obvious that under different initial samples, the fluctuation of the detection accuracy obtained by the network is not large. It benefits from the fact that the massive data of COCO2017 can flexibly adjust the offset. But, it does not mean that the network obtains detection accuracy that are not significantly different at all initial sampled coordinates. To explore again the effect of LDConv with different initial shapes on the network, we explore LDConv with size 5 and with different initial samples for experiments based on YOLOv5n on VisDrone-DET2021. It can be seen in Table~\ref{detect7} that the network obtains different detection accuracy with different initial samples. Therefore, LDConv with different initial sampled shapes has an impact on the performance of the network. Moreover, for specific networks and datasets, it is important to explore LDConv with appropriate initial sampled shapes to improve network performance.

\begin{figure*}[h!]
	\centering
	\includegraphics[trim=0 0 0 0,clip,scale=0.9]{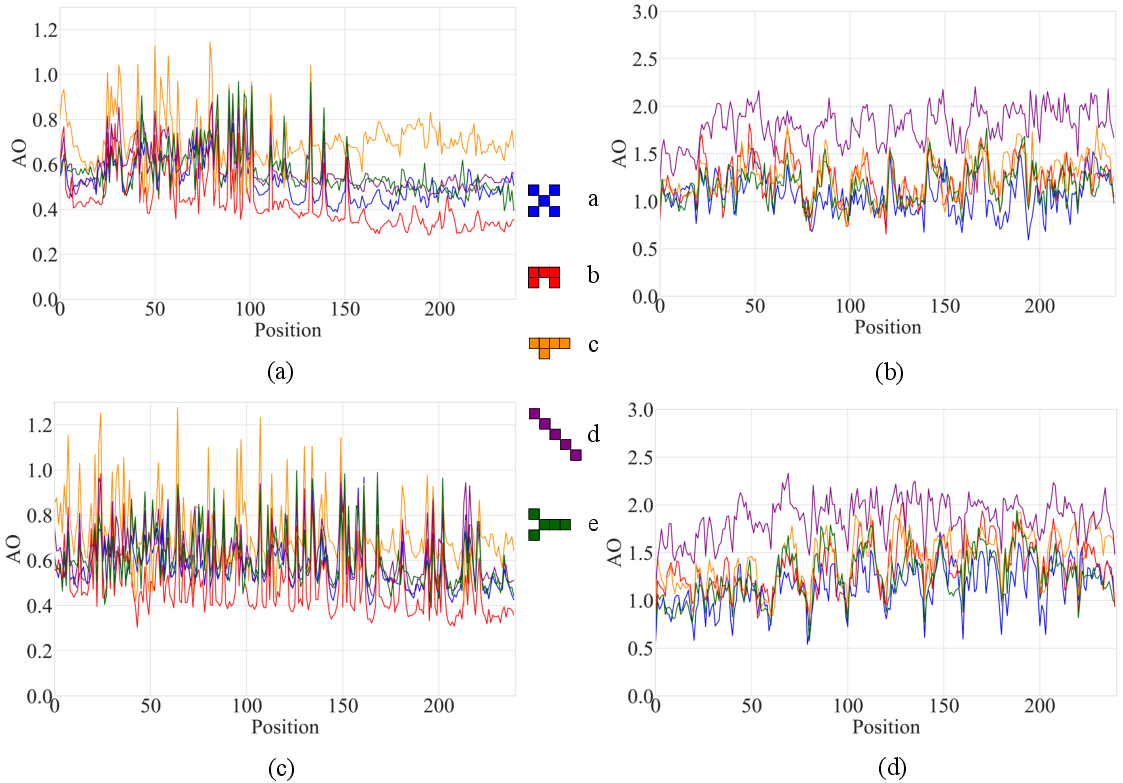}
	\caption{The variation of AO of LDConv for different initial sampled shapes of size 5. It can achieve arbitrary sampled shapes by designing different initial sampled shapes for LDConv.}
	\label{get_pn}
\end{figure*}
\section{Analysis and discussion}
In previous experiments, LDConv with different initial sampled shapes of size 5 was used to evaluate the performance of YOLOv5n. It can be clearly noticed that the network behaves differently under different initial sampled shapes. It suggests that the adjustment ability of offsets is also limited. To measure the change in offset at each given position, we give the definition of the Average Offset, which is defined as follows:
\begin{equation}
\begin{aligned}
AO =(\sum_{i}^{2N}  |Offset_{i}|) / (2N)
\end{aligned}
\end{equation}

AO (Average Offset) measures an average degree of change in the sampled points at each position by summing the offsets, and then taking the average. To observe the change of offsets, we select the trained network and choose the last layer of LDConv to analyze the overall change trend of offsets. For the analysis, we randomly select four images in VisDrone-DET2021 and then visualize the LDConv of size 5, which is initial for different sampled positions. As shown in Figure~\ref{get_pn}, we visualize the degree of change in AO of the offset at each sampled location. The different colors in Figure~\ref{get_pn} represent the change in offsets at each sampled position for different initial samples after training. The color of the line corresponds to the initial sampled shape in the middle.  The different initial sampled shapes in Figure~\ref{get_pn} correspond to the initial sampled shapes in Table~\ref{detect7}. It can be concluded that OA changes less for the blue and red initial sampled shapes in Figure~\ref{get_pn}. It means the red and blue initial samples are more suitable for this dataset than the other initial samples. As in the experiment in Table~\ref{detect7}, it can be seen that the initial sampled shapes corresponding to blue and red obtained better detection accuracy. All the experiments proved that LDConv is able to bring significant performance improvement to the network. Unlike Deformable Conv, LDConv has the flexibility to scale network performance based on size. In all experiments, we explore LDConv with size 5 extensively. Because when training COCO2017 with a large amount of data, we found that when setting the size of LDConv to 5, the training speed is not much different from the original model. Moreover, as the size of LDConv increases, the training time gradually increases.  In the experiments of COCO2017, VOC 7+12, and VisDrone-DET2021, LDConv with size set of 5 gave good results for the network. Of course, the exploration of LDConv for other sizes is possible because the number of parameters that show linear growth and arbitrary sampled shapes bring a wealth of choices for the exploration of LDConv. LDConv can realize convolution operation with arbitrary sizes and arbitrary samples, and can automatically adjust the sampled shape to adapt to the target change by offsets. All experiments demonstrate that LDConv improves network performance and provides richer options for the trade-off between network overhead and performance. Moreover, although LDConv improves the detection accuracy of the model, it sacrifices some detection speed during train. The reason for this is that LDConv uses too much time to access the memory, Therefore in the future, we will design appropriate operators to reduce the memory access time to improve the speed of LDConv.

\section{Conclusion}
It is obvious that in real life as well as in the field of computer vision, the shapes of objects show various variations. The fixed sampled shape of convolutional operation can not adapt to such changes. Although Deformable Conv can flexibly change the sampled shape of convolution with the adjustment of offset, it still has limitations. Therefore, LDConv is proposed in this work, which truly realizes to allow convolution to have arbitrary sampled shapes and sizes, which providing diversity in the choice of convolution kernels. Moreover, the different initial sampled shapes are explored and improve the FasterBlock and GSBottleneck. Although we have designed multiple shapes of sampled coordinates only for LDConv of size 5 in this work, the flexibility of LDConv is that it can target any size of sampled kernel to extract information. Therefore, in the future, we would want to explore LDConv with appropriate sizes and sampled shapes for specific tasks in the field, which will add momentum to the subsequent tasks.

\section*{Declaration of Competing Interest}
The authors declare that they have no known competing financial interests or personal relationships that could have appeared to influence the work reported in this paper.

\bibliographystyle{elsarticle-num}


\end{document}